\documentclass[sn-mathphys]{sn-jnl}

\jyear{2021}%

\theoremstyle{thmstyleone}%
\theoremstyle{thmstyletwo}%

\usepackage{cleveref}

\theoremstyle{thmstylethree}%
\newcommand{\RR}{\mathbb{R}}
\newcommand{\gio}[1]{\textcolor{black}{#1}}
\newcommand{\melo}[1]{\textcolor{black}{#1}}

\begin{document}
\title[DCT-Former: Efficient Self-Attention with Discrete Cosine Transform]{DCT-Former: Efficient Self-Attention with Discrete Cosine Transform}


\author*[1,2]{\fnm{Carmelo} \sur{Scribano}}\email{carmelo.scribano@unimore.it}
\equalcont{These authors contributed equally to this work.}

\author*[1]{\fnm{Giorgia} \sur{Franchini}}\email{giorgia.franchini@unimore.it}
\equalcont{These authors contributed equally to this work.}

\author[1]{\fnm{Marco} \sur{Prato}}\email{marco.prato@unimore.it}
\author[1]{\fnm{Marko} \sur{Bertogna}}\email{marko.bertogna@unimore.it}

\affil[1]{\orgdiv{Department of Physics, Informatics and Mathematics}, \orgname{University of Modena and Reggio Emilia}, \orgaddress{\city{Modena}, \country{Italy}}}

\affil[2]{\orgdiv{Department of Mathematical, Physical and Computer Sciences}, \orgname{University of Parma}, \orgaddress{\city{Parma}, \country{Italy}}}



\abstract{Since their introduction the Trasformer architectures
emerged as the dominating architectures for both natural language processing and, more recently, computer vision
applications. An intrinsic limitation of this family of ``fully-attentive” architectures arises from the computation of the dot-product attention, which grows both in memory consumption and number of operations as $O(n^2)$ where $n$ stands for the input sequence length, thus limiting the applications that require modeling very long sequences. Several approaches have been proposed so far in the literature to mitigate this issue, with varying degrees of success. Our idea takes inspiration from the world of \textit{lossy} data compression (such as the JPEG algorithm) to derive an approximation of the attention module by leveraging the properties of the Discrete Cosine Transform. An extensive section of experiments shows that our method takes up less memory for the same performance, while also drastically reducing inference time. 
Moreover, we assume that the results of our research might serve as a starting point for a broader family of deep neural models with reduced memory footprint. The implementation will be made publicly available at \url{https://github.com/cscribano/DCT-Former-Public}}.




\maketitle

\section{Introduction}\label{sec0}
Transformers are a family of recently introduced Deep Learning (DL) models which leverage the mechanism of dot-product attention to map a sequence of tokens of arbitrary length into a new set of tokens. Thanks to their outstanding performance in a variety of tasks, transformers are nowadays ubiquitous in state-of-the-art techniques that gain any benefit from modeling long-term interactions between elements of a sequence. Another important advantage of transformers is the ability to process sequences of arbitrary length in a single forward pass without incurring the limitations of recurrent approaches: no other standard Machine Learning (ML) or DL methods in the literature have shown this great adaptability so far. In the domain of Natural Language Processing (NLP) transformers are pervasive in any sort of task, such as Machine Translation \cite{vaswani2017attention, devlin2018bert, radford2019language, brown2020language}, text classification, document retrieval, document summarization and several others more. More recently, researchers started to focus on exploiting the benefits of the self-attention mechanism for computer vision tasks \cite{dosovitskiy2020image, carion2020end, Liu_2021_ICCV}, either standalone or applied downstream to a convolutional backbone and even to multimodal problems where the language and visual input needs to be correlated.\\
\noindent Despite the clear benefits that were widely popularized by the recent achievements, the main limitation of this class of models arises from the increase in both memory occupation and computational cost, which grows quadratically with the length of the input sequence. 
\melo{This problematic poses a significant limitation to the application of attention models to process long sequences. The quadratic growth in memory occupation, in particular, imposes an upper-bound on the maximum length on the sequence that can be processed.}
While a multitude of approaches has already been proposed in the literature to mitigate this issue, ideally aiming at making the cost of the attention grow \textit{linearly} with the input's length, the formulation of those solutions is often obscure and poorly interpretable.\\
In this work, we investigate a \melo{method to mitigate the problem of the quadratic dependence on the input's length} through the use of Discrete Cosine Transform (DCT)\cite{ahmed1974discrete}. The DCT, widely used in signal approximation problems and especially in image compression, is well known as the most used linear transform for lossy compression. In our work, we employ DCT to compute an approximation of the real attention and to exploit such compressed representation as a replacement for the full attention. Our methodology, contrarily to other approaches, has a simple formulation and can be clearly interpreted as a mere signal filtering operation. Moreover, the proposed relaxations to the attention's formulation can be experimentally validated against a best-case scenario formulation. We evaluate our methodology both in terms of algorithmic complexity and in terms of performances in a common NLP benchmarking scenario. In particular, we follow the standard approach of pre-training on a large corpus of unlabeled text in an unsupervised fashion and then finetuning on downstream supervised tasks, considering the problem of sentiment classification \cite{maas-etal-2011-learning} as our benchmark. It must be clear by now that the objective is not to propose a new model for language modeling tasks to compete against the state of the art. 
Based on a robust mathematical tool such as DCT, the mathematical treatment is also robust and the potential applications of the method are numerous. \newline

\noindent Our contribution can be summarized as follow:
\begin{enumerate}
    \item We propose a simple yet effective self-attention approximation by leveraging the properties of the DCT.
    \item We experimentally show that our formulation allows for both reduced memory footprint and faster inference, while still being competitive on NLP tasks.
    \item We compare our method against prominent competitors in the literature, showing that our method offers the best trade-off between inference time and model accuracy.
\end{enumerate}

Structure of the paper: in \Cref{sec1} we introduce the main mathematical models and their formulations. In \Cref{sec2} we present an overview of the prominent methods in the literature whose purpose is to reduce the computational complexity of the mechanism of self-attention. Additionally, we present a small overview of methods that leverage Fourier-affine transforms in deep learning models. Subsequently, in \Cref{sec3} we introduce our proposed methodology to approximate the self-attention with \textit{quasi-}linear cost. Finally, in \Cref{sec4} we detail our experimental setup and we report and discuss an experimental evaluation of the proposed methodology in multiple NLP applications.

\section{Background}\label{sec1}

\subsection{Neural Networks and Sequential models}\label{sec1:1}

Feed-Forward (FF) Artificial Neural Networks (ANN) are the simplest kind of DL model \cite{GoodBengCour16}. A standard FF network is a nonlinear function \(\Psi: \RR^d \rightarrow \RR^k\), which maps an input \(x \in \RR^d\) into an output \(y\in \RR^k\). In general, the function \(\Psi\) takes the form of a stack of Fully-Connected \textit{layers}. Each layer is defined by a weights matrix \(W_i \in \RR^{k_{i-1}\times k_{i}}\) plus a scalar bias term \(b_{i-1}\) with $i=1, \dots, L$, where $L$ is the number of layers and $k_0=d, k_L=k$. Each layer is also characterized by a nonlinear activation function \(\sigma_i(\cdot)\). The recursive formula takes the form:
\begin{equation*}
y_i = \sigma_i(W_i^Ty_{i-1}+b_i)
\end{equation*}
\noindent where $y_0=x$ and $y_L=y$.\\

\noindent On the other hand, the total function can be written as:
\begin{equation*}
    \Psi(x)=\sigma_L(W_L^T\sigma_{L-1}(W_{L-1}^T\dots \sigma_0(W_0^Tx+b_0))+b_L).
\end{equation*}

\noindent An important limitation of the FF model is the inability to operate with inputs of non-fixed length, which would be desirable to work with sentences in natural language and other kinds of sequential data points. In NLP, an input sentence is usually split into a set of tokens, which represents individual dictionary indices, and each token is mapped to an \textit{embedding} of fixed dimension. For convenience here, we use the terms token and embedding interchangeably, since it persists a $1:1$ mapping.\\
In contrast, Recurrent Neural Networks (RNN) \cite{hochreiter1997long, cho2014properties} arise expressively for the management of sequences, whether they are sequence connected by a temporal component (e.g., time series) or meaning (e.g., a sentence). An RNN processes a sequence of inputs \(X\in \RR^{n\times d}\) by feeding sequentially each row element of $X$ \(x_i\in \RR^d\),  referred to as a \textit{token}, to a stack of recurrent cells. In this case, the recurrent formula is:
\begin{equation*}
y_i(t) = \sigma_i(W_i^Ty_{i-1}+\overline{W}_i^Ty_i(t-1)+b_i)\quad t=(0,1,...,n)
\end{equation*}
where $\overline{W} \in \RR^{k_{i-1}\times k_{i}}$ is a weight matrix trained during the epochs.\\
This implies that the \textit{i}-th activation for the \textit{t}-th token will also depend on the activation produced by the input provided at a previous timestep, effectively allowing to consider \(y_i(t)\) as a fixed size latent representation of the whole input sequence up to \textit{t}. The sequential nature of the recurrent cell however poses a severe performance bottleneck by requiring the tokens to be fed to the model one after the other. Such throughput limitation, together with several other problems in terms of expressive capacity,
are among the reasons that lead to the introduction of attentive and ultimately fully-attentive models.



\subsection{Transformers}\label{sec1:2}

\begin{figure*}[h!]
    \centering
    \includegraphics[clip,  trim=0cm 0cm 0cm 0cm, width=0.9\textwidth]{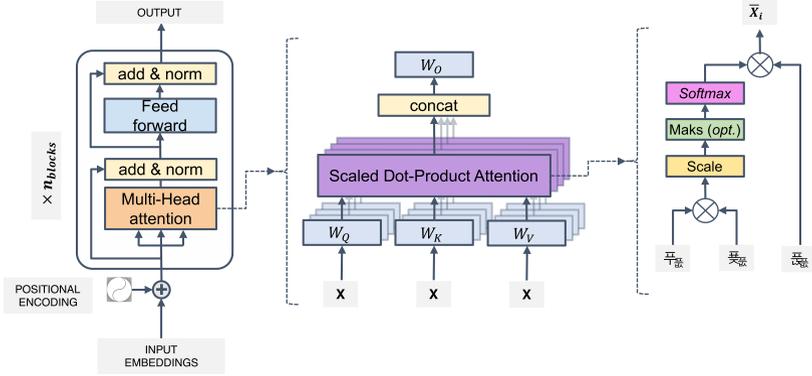}
    \caption{Overview of the general architecture of the standard Transformer Encoder.}
    \label{fig:transformer_arch}
\end{figure*}

Transformers \cite{vaswani2017attention} represent the current state-of-the-art in DL models for sequence modeling tasks. This family of architectures 
replaces the recursion mechanism of RRNs with the introduction of the mechanism of \textit{self-attention} to effectively process sequences of arbitrary length in a single forward operation. As in \Cref{fig:transformer_arch}, a standard transformer is made of $\mathbf{n_{blocks}}$ identical blocks, each composed of two sub-blocks: a self-attention module and a feed-forward layer, each one followed by a layer normalization \cite{ba2016layer} operation and a residual connection.

\paragraph{Self attention}
Given a set of tokens \(X\in \RR^{n\times d}\), the self-attention mechanism produces a similar set of tokens \(\mathcal{X}\in \RR^{n\times d}\), where each new row token element of $\mathcal{X}$ 
is obtained as a weighted average of the whole original set \(X\).\\

The resulting weights of the attention mechanism represent the affinity degree between pairs of tokens. Such affinity is computed by first projecting, where with the term projection we mean a simple matrix multiplication, \(X\) onto a set of \textit{Queries} \(Q\in \RR^{n\times d_q}\), a set of \textit{Keys} \(K\in \RR^{n\times d_k}\) and a set of \textit{Values} \(V\in \RR^{n\times d_v}\), with three distinct projection matrices \(W_Q \in \RR^{d\times d_q}\) , \(W_K \in \RR^{d\times d_k}\), and \(W_V \in \RR^{d\times d_v}\). The three projections are shown below:
\begin{equation}\label{eq:atn_qkv}
    Q = XW_Q,\quad K=XW_K,\quad V = XW_V
\end{equation}
The dot-product between \(Q\) and \(K^T\) (with $d_q=d_k$ and $d_v=d$ in self-attention) produces an \textit{Energy} score between pairs of tokens, which is then normalized and fed to a nonlinear \textit{softmax} \cite{bridle1989training} operation to obtain the final weights matrix. In this case the \textit{softmax} operation is applied row-wise.
\begin{equation}\label{eq:energy}
E(Q,K) = softmax\left(\frac{QK^T}{\sqrt{d_q}}\right)
\end{equation}
\noindent The energy is finally multiplied by \(V\) in order to produce the final attention output:
\begin{equation}\label{eq:attention_full}
Atn(X) = E(Q,K)V = softmax\left(\frac{XW_Q(XW_K)^T}{\sqrt{d_q}}\right)XW_V
\end{equation}
In almost any transformer implementation a number {$\mathbf{n_{heads}}$} of self-attention heads, each with its own set of projection matrices \(W_{Q,K,V}^j\), with $j=1,\dots,n_{heads}$, are applied in parallel, defining the Multi heads Self Attention (MhSA). The output of the \textbf{multi-head} attention is obtained as a concatenation of the results of the individual attention heads, usually followed by an additional projection layer \(W_O \in \RR^{md_v \times d}\).
\begin{equation*}
\mathcal{X}=MhSA(X) = [Atn_1(X)\oplus Atn_2(X)\oplus ....\oplus Atn_M(X)]W_O
\end{equation*}
For the sake of simplicity, from now on we can ignore the multi-head aspect of the transformers attentions, since the problem that we investigate is not dependent on the number of attention heads but is related to a single attention term.

\paragraph{Quadratic complexity of Attention}\label{paragraph:ccatn}
It is clear from \eqref{eq:energy} that being \(n\) the sequence length, the complexity of calculating the attention's weight matrix \(E \in \RR^{n\times n} \) is \(O(n^2)\) in both memory and time, which limits significantly the applicability of the self-attention mechanism for very long input sequences.
To overcome the limitations of the quadratic dependence, several options have been already proposed in the literature, some of which are discussed in \cref{sec3:2}.

\subsection{Transformer based Language modeling}\label{sec1:3}

Given the property of the self-attention mechanism, since their introduction, transformers have been popularized as powerful language modelers. However, transformer-based language models are known to be extraordinarily hard to train by relying only on labeled data for supervised tasks. For this reason, the scheme of adopting a pre-training strategy, already popular in previous language modeling techniques \cite{pennington2014glove, peters2017semi}, has become of great importance for transformers based modeling. Pre-trained transformers can be then effectively fine-tuned for downstream supervised tasks, usually with little to none architectural changes.\\
Among the considerable variety of pre-trained transformer models, BERT \cite{devlin2018bert} and its derivatives \cite{lan2019albert, clark2020electra, he2020deberta, liu2019roberta, iandola2020squeezebert} have become the de facto standard for deep language modeling. The strength of this model comes from the \textit{bidirectional} pretraining strategy, which leverages a huge amount of unlabeled text in an unsupervised fashion. From an architectural standpoint, BERT simply employs the original transformer architecture adding a WordPiece tokenizer \cite{wu2016google} to split an input sentence in a sequence of dictionary entries, which are then mapped to token embeddings. The unsupervised pre-training is carried by simultaneously optimizing for two tasks: 
\begin{itemize}
\item Masked Language Modeling (\textbf{MLM}), where a percentage of the input tokens is masked at random, by replacing those with a placeholder \texttt{[MASK]} token, and then asking the model to predict back the masked tokens.
\item Next Sentence Prediction (\textbf{NSP}) task, where a pair of sentences (sentence \texttt{A} and sentence \texttt{B}) are fed together to the model, divided by a separation token \texttt{[SEP]}, and the model is tasked to classify whether the sentence \texttt{B} is the actual next sentence that follows \texttt{A} or is a random sentence from the training corpus. 
\end{itemize}
The training corpus is obtained by combining BooksCorpus \cite{zhu2015aligning} and English Wikipedia in order to obtain over 3,5M words of document-level corpus which include long sequences of sentence-level text required for the pre-training objectives.
A major downfall of the transformers pre-training is the very large computational power required to achieve state-of-the art performance, 
with a proper training easily approaching costs in the tens of thousands dollars \cite{sharir2020cost} (based on the current cloud GPU prices).
For our experimental validation we trained a BERT\textit{-like} model following the training recipe detailed in \cite{izsak2021train}, while the language modeling ability of such a model cannot be compared with a full pre-training BERT, but it is instead perfectly suited to demonstrate the advantage of our approximated attention in a fair comparison scenario. In \Cref{sec4} we detail the experimental setup and the adopted training scheme.


\subsection{Discrete Cosine Transform}\label{sec1:4}
The DCT \cite{ahmed1974discrete} is a Fourier-related transform which expresses a finite sequence of elements (a discrete signal) in terms of a sum of cosine functions at different frequencies. Most noticeably, the DCT is both discrete and, contrary to the Discrete Fourier Transform (DFT), real-valued. DCT is invertible, with the inverse function denoted as IDCT, and enjoys the properties of \textit{energy compaction}, concentrating the energy of the signal in few coefficients, and \textit{decorrelation}, since the coefficients are uncorrelated to each other. Thanks to those properties, DCT is heavily used as a transformation mechanism in signal processing, and especially in \textit{lossy} data compression algorithms such as JPEG (images) \cite{raid2014jpeg}, MPEG (video), and MPEG Layer III or MP3 (digital audio).  
There are several variants of DCT, the most common, also used in this work, is the type-II DCT \cite{SHAO20081553}, which was also the first version of DCT.\\ 
Given a finite length sequence of $N$ real valued elements $\hat{x} \in \RR^{N\times 1}$, the Type-II DCT is a sequence $\hat{X}$ of the same length defined as:
\begin{equation}\label{eq:dct}
    \begin{gathered}
    \hat{X}_K = \alpha_k \sum_{n=0}^{N-1} x_n \cos\left(\frac{\pi(2n+1)k}{2N}\right) \quad \text{for }  k=0,1,...,N-1\\
    \text{where }\alpha_k=
        \begin{cases}
          \sqrt{1/N} & \text{if $k=0$}\\
          \sqrt{2/N} & \text{if $k\ne0$}\\
        \end{cases}
    \end{gathered}
\end{equation}
Since the DCT is a linear transformation, \eqref{eq:dct} can be conveniently expressed in terms of a dot-product operation between the sequence $x$ and a transformation matrix $D\in\RR^{N\times N}$. Formally, $\hat{X}=DCT(\hat{x}) = D\hat{x}$, where:
\begin{equation}\label{eq:dct_matrix}
D_{n,k} = \alpha_k \cos\left(\frac{\pi(2n+1)k}{2N}\right)
\end{equation}
\noindent Due to the normalization term $\alpha$, the matrix $D$ is \textit{orthogonal}, which makes possible to express the inverse transform as \(D^{-1} = D^T\), hence: \(IDCT(\hat{X}) = D^T\hat{X}=\hat{x}\), thus avoiding the high computational cost of the inverse calculation.\\
Generally, when we speak about lossy compression algorithms, we first compute the computing of the DCT coefficients of a signal, and then keep only a handful of the most relevant values. A simple way of computing a low-frequency approximation would be to define a matrix $\overline{D} \in \RR^{M\times N}$ with $M<N$ by keeping only $M$ rows of the transformation matrix $D$. $\overline{D}$ can be used to obtain a compressed representation $\overline{x} \in \RR^{M\times 1}$ by computing the forward DCT, then a lossy reconstruction of the original $\hat{x}$ is obtainable with the inverse transform.\\
When we are dealing with a transformation that is performed by both rows and columns, we can generalize the observations made before by using 2D-DCT. In this case, given a finite length sequence of $N \times N$ real elements $\hat{x} \in \RR^{N \times N}$, the 2D-DCT can be computed with the formula $\hat{X}=DCT(\hat{x})=D\hat{x}D^T$. Using the same methodology as described above, we can generalize the compression procedure to the 2-dimensional case. 

\section{Related Works}\label{sec2}

\subsection{Efficient Attention Heads}\label{sec2:1}

As briefly mentioned in the introduction, the quadratic complexity of the attention is a well-studied issue in the deep learning community. A variety of solutions have so far already been proposed, which can be roughly categorized in three classes: (i)~methods that try to approximate or factorize the attention as defined in the original formulation \cite{wang2020linformer, choromanski2020rethinking, kitaev2019reformer, xiong2021nystromformer, lu2021soft, ren2021combiner,nguyen2021fmmformer} (ii)~methods that reformulate the definition of attention (e.g., by introducing locality constraints) to avoid the complexity bottleneck \cite{wu2021fastformer,jaszczur2021sparse,beltagy2020longformer,tay2021synthesizer,zhu2021long,chen2021scatterbrain} (iii)~contributions which entirely remove the self-attention, usually by proposing an alternative paradigm \cite{tolstikhin2021mlp, lee2021fnet, you2020hard}. Our methodology clearly falls in the first category, therefore hereafter we provide a brief description of our principal competitors, a few of which will be used for comparison in the experimental \Cref{sec4}.

\melo{\paragraph{Attention Matrix Reduction}}
\noindent Reformer \cite{kitaev2019reformer} achieves a complexity of $O(nlog(n))$ by reducing the number of operations in the computation of $softmax(QK^T)$ \eqref{eq:energy} introducing a local-sensitivity-hashing (LSH) mechanism. Their methodology is based on the observation that large values dominate the output of the $softmax$ operation, hence they claim to be sufficient to only compute the largest values of the $QK^T$ product. In Linformer \cite{wang2020linformer} the authors introduce a set of learnable linear projection matrices $\overline{\epsilon}_{i}  (i=1,..,n_{heads})$ to project $Q$ and $V$ in a lower dimensional space, justifying this approach with the empirical observation of the attention matrix being low-rank.
Performers \cite{choromanski2020rethinking} introduce a kernelizable attention mechanism (FAVOR+) to approximate the softmax attention with a complexity of $O(n)$. More recently, Nystr\"omformer exploited the usage of the Nystr\"om approximation which is commonly used in kernel methods to approximate the Gram matrix (positive semi-definite) with a low rank matrix. To avoid computing the full attention, the authors exploit a relaxation of the Nystr\"om method by individually computing the softmax operation of the three decomposition sub-matrices before the dot product operation.\\
\noindent SOFT \cite{lu2021soft} builds on top of \cite{xiong2021nystromformer} by replacing the dot-product operation with a Gaussian kernel, thus entirely removing the softmax operation from the formulation allowing, for a proper application of the Nystr\"om method. Moreover, they propose a Newton-Raphson based method to approximate the pseudoinverse operation, in contrast with the less efficient Moore-Penrose pseudoinverse used in \cite{xiong2021nystromformer}.

\melo{
\paragraph{Differences with Model Compression}
Some readers might be familiar with some popular techniques to reduce inference cost of generic deep learning models. Among those, quantization techniques \cite{jacob2018quantization} rely on reduced precision arithmetic (either 8-bits integers or 16-bits floating point), pruning \cite{NIPS2015_ae0eb3ee} remove less important weights or nodes from the network, and knowledge distillation \cite{hinton2015distilling} is a technique to transfer the knowledge of a large model in a smaller one.\\
The formulation detailed in this manuscript, as well as the competitors previously introduced, are not related to these compression strategies. Formulations for efficient attention focus on mitigating the issue of the quatratic complexity of the dot-product attention, while compression strategies are aimed exclusively at reducing inference times and are often tailored to the particular capabilities of the hardware used for inference \cite{vanhoucke2011improving}.
}

\subsection{Frequencies domain}
\gio{
In the frequency domain, a matrix which represents a digital image is converted from spatial to frequency domain. The Fast Fourier Transform is an efficient method used to convert the spatial to the frequency domain. In this paper, DCT was specifically chosen to transform attention matrix information into frequencies because of some of its characteristics: DCT operates in the real field like images, its compression capability has been demonstrated and widely used in the literature, and its matrix formulation makes its computation and the computation of its inverse particularly efficient on parallel architectures. Regarding the latter point, many works have dealt with FFT parallelization, as e.g. \cite{ZHOU20071402} in which the authors propose a novel and hardware-efficient architecture for power-of-two FFT processors.\\
In the signal processing literature there is extensive use of the DCT/FFT in "learning" problems. For example in \cite{SHAO20081553} the authors proposed an efficient and flexible dictionary structure for sparse and redundant signal representation and they demonstrated the advantages of the proposed structure for 3-D image denoising. On the other hand, in \cite{7938674} orthogonal and nonorthogonal dictionaries are factorized as a product of a few basic transformations to balance data representation performance and computational complexity. Also in \cite{7178579} the authors work with dictionary learning, with the aim of finding a frame (called dictionary) in which some training data admits a sparse representation. The approach is demonstrated experimentally both with a factorization of the Hadamard matrix and on image denoising.
}
\subsection{Neural Networks in the frequency domain}\label{sec2:2}
To the best of our knowledge, only few works have so far exploited Fourier-related transforms in the DL domain. A remarkable contribution is the recent F-Net \cite{lee2021fnet}, which entirely replaces the transformer's self-attention with a two-dimensional Discrete Fourier Transform operation. While this might sound similar to our methodology (\Cref{sec3}), it is entirely different in the formulation, since their method does not represent an approximation for the dot-product operation but rather a complete substitute. Previously \cite{gueguen2018faster} proposed to operate a Convolutional Neural Network (CNN) on the DCT coefficients of a JPEG compressed image to avoid the need to run the full JPEG decoding algorithm. Several other contributions, such as \cite{dziedzic2019band, rajesh2019dct, xu2020learning, dos2020good}, explored similar concepts for computer vision problems with varying degrees of success.

\section{Proposed Methodology}\label{sec3}

\subsection{A Naive Solution}\label{sec3:1}
We recall that the goal of our investigation is to exploit the DCT introduced in \Cref{sec1:4} to define an approximation method which avoids a quadratic growth of the attention matrix in \eqref{eq:energy} with the input sequence length $n$ for an input
$X \in \RR^{n\times d}$.\\
Given the three $(n\times d)$ matrices $Q$, $K$ and $V$ defined in \eqref{eq:atn_qkv}, all functions of input $X$, a straightforward solution is to individually obtain three compressed representations $\overline{Q}$, $\overline{K}$ and $\overline{V}$ each of length $\overline{n} << n$ by computing the DCT of each matrix over the dimension $n$ and retaining only $\overline{n}$ DCT coefficients. For ease of understanding we can express the forward DCT relying on the matrix formulation of \eqref{eq:dct_matrix}, hence a transformation matrix $\overline{D} \in \RR^{\overline{n}\times n}$ can be easily obtained from the definition to compute the required $\overline{n}$ DCT coefficients.\\
Denoting the transformation matrix as $\overline{D}$ we formulate:
\begin{equation}\label{eq:dct_dkv}
    \overline{Q} = \overline{D}Q,\quad \overline{K} = \overline{D}K,\quad \overline{V} = \overline{D}V
\end{equation}
by substituting in \eqref{eq:energy} we obtain:
\begin{equation*}
\overline{E}(\overline{Q},\overline{K}) = softmax\left(\frac{(\overline{D}Q)   (K^T\overline{D}^T)}{\sqrt{d}}\right)
\end{equation*}
If we consider the numerator inside the softmax operator is clear to see that:
\begin{equation*}
    \overline{D}(Q   K^T)\overline{D}^T = DCT_{2D}(QK^T)
\end{equation*}
By leveraging the associative property of the dot-product, $\overline{E} \in \RR^{\overline{n}\times \overline{n}}$ is obtained without explicitly computing the original $E \in \RR^{n\times n}$. Going forward, the compressed attention output is computed by multiplying with $\overline{V}$:
\begin{equation}\label{eq:dct_atn}
    \overline{Atn}(X) = \overline{E}(\overline{Q},\overline{K})\overline{V}
\end{equation}
And finally, the resulting approximated attention $\widetilde{Atn}(X)$ is obtained with an inverse DCT:
\begin{equation}\label{eq:idct_atn}
    \widetilde{Atn}(X) = IDCT(\overline{Atn}(X)) = (\overline{D})^T\overline{Atn}(X)
\end{equation}
From \eqref{eq:dct_atn}, \eqref{eq:idct_atn}:
\begin{equation*}
    \widetilde{Atn}(X) = (\overline{D})^T[\overline{E}(\overline{Q},\overline{K})]\overline{D}V = IDCT_{2D}(\overline{E})V
\end{equation*}
To reiterate, our approximated attention grows in memory and complexity with $O(\overline{n}^2)$, by picking an $\overline{n}$ small enough is possible to approach a linear growth with the original input length $n$. A clear relaxation in our method is that we are in effect leveraging:
\begin{equation}\label{eq:softmax_relax}
    \widetilde{Atn}(X) = IDCT_{2D}(softmax(DCT_{2D}(QK^T)))V
\end{equation}
where the normalization term $\sqrt{d_{e}}$ is omitted. Clearly $softmax(DCT(x)) \neq DCT(softmax(x))$, hence by computing the Inverse DCT we are implicitly introducing a relaxation. Similar relaxations involving the softmax function have already been proposed in \cite{wang2020linformer} and \cite{xiong2021nystromformer}, in \Cref{sec4:3} we discuss in detail its implications and define a strategy to experimental evaluate the performance degradation caused by its utilization.

\subsection{A More efficient formulation}\label{sec3:2}
A first improvement that we can introduce to make our formulation more efficient from a computational standpoint is to avoid the calculation of three distinct forward DCT transforms as in \eqref{eq:dct_dkv}. Recalling the formulation for $Q, K$ and $V$ in \eqref{eq:atn_qkv}, we can save on computation by computing only the DCT of $X$ ($\overline{X} = \overline{D}X$) and then utilize the compressed $\overline{X}$ in place of $X$ in the attention formulation of \eqref{eq:attention_full}. This can be easily proven to be equivalent to the approximated attention defined in \eqref{eq:dct_atn}. Our formulation can be then formalized as in \Cref{algo1}.\\
\begin{algorithm}[h]
\caption{Efficient attention with DCT}\label{algo1}
\textbf{Input} $X \in \RR^{n\times d}$\\
\textbf{Output}  $\widetilde{X} \approx Atn(X) $
\begin{algorithmic}[1]
\Require $\overline{D}\in \RR^{\overline{n}\times n}$
\State $\overline{X}=DCT(X)=\overline{D}X$
\State $\overline{Q} = \overline{X}W_Q,\quad \overline{K}=\overline{X}W_K,\quad \overline{V} = \overline{X}W_V$
\State $Atn(\overline{Q},\overline{K},\overline{V}) = E(\overline{Q},\overline{K})\overline{V} = softmax\left(\frac{\overline{Q}\overline{K}^T}{\sqrt{d}}\right)\overline{V}$
\State $\widetilde{X} = \overline{D}^T\left[Atn(\overline{Q},\overline{K},\overline{V})\right]$\\
\Return $\widetilde{X}$
\end{algorithmic}
\end{algorithm}
From the efficiency standpoint, the choice of the matrix formulation to compute the DCT is optimal: a single $\overline{D}$ can be precomputed, memorized and shared across all the attention modules of the transformer architecture of choice. This for example in stark contrast with \cite{wang2020linformer} where each attention head requires its own learnable projection matrix, resulting in a total of $(n_{heads}*n_{blocks})$ matrices to be stored in memory.
Moreover, relying on a known linear transformation matrix has its own set of advantages: (i)~it reduces the total number of trainable parameters, making for a lighter and more efficient training (ii)~learning a transformation matrix $E \in \RR^{n\times \overline{n}}$ as in \cite{wang2020linformer} implies that the input sequence length must be exactly $n$, taking away the option of model input of arbitrary lengths. When using the DCT instead we can easily recompute $\overline{D}$, or we can exploit an algorithm for fast cosine transform without explicitly relying on $\overline{D}$. For the latter, in the fine-tuning experiments (\Cref{sec4:2})  we employed Makhoul's algorithm \cite{makhoul1980fast}, which leverages the Fast Fourier Transform (FFT) to efficiently compute the DCT of a $N-$point real valued signal.\\
To conclude, among the different transforms available in the literature, DCT was chosen because it is an efficient way of compressing information, it can be expressed as a matrix product, its inverse calculation is also linear and operates in the real numbers field.


\subsection{The curse of nonlinear softmax}\label{sec3:3}

As introduced in \Cref{sec4:1}, a significant relaxation exploited by our formulation to compute the inverse DCT of the result of a nonlinear function $softmax(x): \RR^n \rightarrow \RR^n$ applied to the result of the forward DCT, as from \eqref{eq:softmax_relax}.\\
For the sake of completeness, we recall that the softmax function is defined as:
\begin{equation*}
    softmax(x)_{i} = \frac{e^{x_{i}}}{\sum_{j=0}^{n}{e^{x_j}}} \quad i=0,1,...,n
\end{equation*}
This function is commonly used in the deep learning domain to highlight larger values and hide the ones significantly smaller than the maximum; moreover, it constrains the output of a layer to sum to $1$ and returns values between $0$ and $1$.\\
Ideally, to avoid our relaxation we would need a function $\overline{s}$ such that:
\begin{equation*}
    \overline{s}(DCT(x)) = DCT(softmax(x)).
\end{equation*}
This function is trivially $\overline{s}(DCT(x))=\overline{s}(\hat{x})=D(softmax({D}^T\hat{x}))$ that is very unsuitable since it implies passing through a higher-dimensional space, which is exactly what we want to avoid with the proposed method. With our relaxation, we can instead avoid the computation of the matrix $E\in\RR^{n\times n}$.\\
\noindent When leveraging our formulation, we introduce two potential sources of error when compared to the standard attention definition:  (i)~an \textit{approximation} error induced by the lossy compression using $\overline{n}<n$ DCT coefficients, and (ii)~a \textit{relaxation} error induced by the usage of the softmax relation above mentioned. The approximation error is intrinsically in the definition of lossy data compression, the relaxation error needs instead to be carefully evaluated in order to prove our methodology to be mathematically worthy.

\begin{algorithm}[h!]
\caption{Evaluation of DCT-induced error}\label{algo2}
\textbf{Input} $X \in \RR^{n\times d}$\\
\textbf{Output}  $\widetilde{X} \approx Atn(X) $
\begin{algorithmic}[1]
\Require $\overline{D}\in
\RR^{\overline{n}\times n}$
\State $Q = XW_Q,\quad K=XW_K,\quad V=XW_V$
\State $E \Leftarrow  E(Q,K) = softmax\left(\frac{QK^T}{\sqrt{d}}\right)$\label{op0}
\State $\overline{E} = \overline{D}E\overline{D}^T \in\RR^{\overline{n}\times \overline{n}}$ \vspace{2.0pt}
\State $\widetilde{E} = \overline{D}^TE\overline{D} \in\RR^{n\times n}$\vspace{2.0pt}
\State $\widetilde{X} = \widetilde{E}V$ \\
\Return $\widetilde{X}$
\end{algorithmic}
\end{algorithm}
We devise a simple yet effective strategy to experimentally evaluate the contribution of the softmax relaxation on the overall error degree: the full matrix $E\in\RR^{n\times n}$ is explicitly obtained as in \eqref{eq:energy}, then its forward and inverse DCT are computed to obtain a lossy reconstruction $\widetilde{E}\in\RR^{n\times n}$ which is then used in the following steps to obtain the attention output.\\
With this setup, the relaxed formulation of \eqref{eq:softmax_relax} is never used, hence only the approximation error is added: a simple way of quantifying the relaxation error is to compare the experimental results obtained with this formulation with those of the efficient attention formalized by \Cref{algo1}. It is worth clarifying that the above setup it is only intended for evaluation and comparison, since the quadratic attention is explicitly computed in \Cref{op0} of \Cref{algo2} there would not be any benefit in using this formulation in a real use scenario.


\section{Experimental Evaluation}\label{sec4}

\subsection{Experimental Setup}\label{sec4:1}
Our experimental setup follows the transfer learning scheme common in NLP: first the model is trained on a large dataset of unlabeled corpus data, then we finetune the model on a downstream supervised task. In subsection \ref{sec4:3} we report the results both on the pretrain and the downstream task, while in subsection \ref{sec4:2} we present the results in terms of inference speed and memory occupation.

\paragraph{Model Architecture}
The transformer architecture adopted for our experiments is inspired by $\mathbf{BERT_{small}}$ introduced in \cite{turc2019well}. The model architecture follows the same structure of the original transformer \cite{vaswani2017attention} while only using $n_{blocks}=4$ instead of the 12 of $BERT_{base}$ in order to keep a reasonable memory footprint even when training with the standard attention head. Each multi-head attention uses 8 heads, the embedding dimension $d$ is 512 and the hidden dimension of the feed-forward layer is 2048. For the input tokenization, we employed the same pretrained WordPiece tokenized used in BERT, leveraging the implementation \textit{``bert-base-uncased"} provided by the Transformers library \cite{wolf-etal-2020-transformers}. 

\paragraph{Pretraining}
For our evaluation, we base our workflow on the pipeline proposed \cite{izsak2021train}, which combines several techniques to train a BERT-style language model with a reasonable computational budget. Following their setup, we optimize only for the masked-language model (MLM) task with a sparse token prediction head \cite{liu2019roberta}, not using the next sequence prediction (NSP) objective, but we used only English Wikipedia text as training corpus. To maximize the training throughput 10 masked copies of the dataset are precomputed, with a masking probability of $0.1$. Moreover the maximum sequence length $n$ is limited to $128$ tokens to allow for larger batch sizes. 
On the optimization side, we mostly followed the same setup using the optimizer AdamW \cite{loshchilov2017decoupled} with $(\beta_{1}=0.9, \beta_{2}=0.98, \epsilon=1e{-6}$) and weight decay of $0.01$. To allow for an unbiased comparison of models with vastly different training speeds, we discarded the fixed time-budged scheduler from the training recipe, instead we fixed the total number of optimization steps to $100k$ and linearly increased the learning rate from $0$ to the Peak-lr with a warm-up proportion of $0.06$, then applied a linear decay for the remainder of the steps. The peak learning rate (LR) is fixed to $1e{-3}$ and the minibatch size to $4096$, obtained with two gradient accumulation steps.\\
From an implementation standpoint the optimization engine DeepSpeed \cite{rasley2020deepspeed} is used with mixed precision training provided by the APEX\footnote{\url{https://github.com/nvidia/apex}} backbone. To avoid potential interferences with the efficient attention formulation, we avoided using fused linear-activation-bias layers and APEX LayerNorm implementation, which are commonly used to speedup training. All our experiments are trained on two 32GB Nvidia V100 GPUs, leveraging model-level parallelism.
\paragraph{Finetuning}
In the spirit of keeping the experimental setup simple and understandable we opted to evaluate our model on the downstream task of sentiment classification of IMDb movies reviews \cite{maas-etal-2011-learning}. This dataset consists of 50.000 movie reviews in plain English text, evenly split between train and test. Each review is manually labeled for sentiment classification as positive or negative depending on the writer's liking of the movie, positive and negative labels are distributed with a $0.5$ ratio both in the test and train splits, making for a perfectly balanced classification task. The sequences of the training set are in average $298$ tokens long (\textit{min.} 13, \textit{max.} 3055), to save memory during training we cap the maximum sequence length to 1024 tokens, truncating the longer sequences. \\ 
To finetune the model, the MLM head used for pre-training is replaced by a classification head. Only the first token $C\in\RR^{d}$ is kept from the transformer's output $\mathcal{X}\in \RR^{n\times d}$. $C$ corresponds to the special \texttt{[CLS]} token, which is added to the input. $C$ is then fed to two feed-forward layers with a $tanh(\cdot) : \RR \rightarrow \RR$ activation function to produce the binary classification output. The model is optimized with a Binary CrossEntropy objective function, as for the pretraining we use the optimizer AdamW, but the learning rate is fixed to $1e{-5}$: in total we train each model for $10$ epochs with a batch size of $64$ with no gradient accumulation steps.

\subsection{Evaluation}\label{sec4:2}

Reducing memory footprint and computational cost is the main objective of our work, therefore hereafter we provide detailed results on the requirements of our model and compare them against the main competitors in the literature. We compare our attention head against the original (\textit{Vanilla} \cite{vaswani2017attention}) transformer implementation as well as Linformer, Nystr\"omformer and Performer.

\paragraph{Model Inference results}
For a fair comparison we used for all the tests our transformer model defined in Section \ref{sec4:1}, replacing only the attention head. We tested with randomly generated sequences of length $n \in \{128, 512, 1024, 4096\}$ adapting the batch size accordingly to fit the model in memory: to adjust for the non fixed batch size we normalize both the inference time and the memory occupation for the current batch size. \melo{All the measurements are taken accounting only for the forward propagation.}\\

\begin{table}[h]
\begin{center}
\label{Table1}
\caption{Inference performance of our efficient attention compared to other attention heads}
\begin{tabular}{ccccccccc} 
\toprule
                & \multicolumn{8}{c}{Sequence length (N) - Batch size (BS)}                                                                    \\ 
\cmidrule{2-9}
Attention Head  & \multicolumn{2}{c}{128 - 256} & \multicolumn{2}{c}{512 - 32} & \multicolumn{2}{c}{1024 - 16} & \multicolumn{2}{c}{4096 - 1}  \\ 
\cmidrule{2-9}
                & MB    & ms                    & MB     & ms                  & MB     & ms                   & MB     & ms                   \\ 
\midrule
Vanilla & 5.1 & 0.391 & 28.75 & 1.99 & 89.37 & 5.03 & 1250.0 & 45.6 \\
DCT-0.25 & 4.55 & \textbf{0.312} & 22.62 & \textbf{1.34} & \textbf{44.5} & \textbf{2.85} & \textbf{326.0} & \textbf{15.75} \\
\cmidrule(lr){1-9}
Linformer-0.125 & \textbf{4.23} & 0.374 & \textbf{21.0} & 1.62 & 46.75 & 3.52 & 612.0 & 19.3 \\
Nystr\"om-0.125 & 4.73 & 0.41 & 24.87 & 1.83 & 55.5 & 4.18 & 488.0 & 47.71 \\
Performer-0.125 & 4.91 & 0,425 & 23.87 & 1.89 & 59.5 & 4.2 & 548.0 & 28.93 \\ 
\bottomrule
\end{tabular}
\end{center}
\footnotetext{The results are expressed in terms of peak memory occupation (Megabytes, MB) and inference time (milliseconds, ms). All the models are evaluated on a single Nvidia 2080Ti}
\end{table}

In Table \ref{Table1} we adopt the notation $\mathbf{\{Model\}-\{scale\}}$ where \textbf{scale} indicates the (fixed) ratio of the input sequence length used to instantiate the efficient attention: for our method it defines the number of DCT coefficients, for Linformer the dimension of the learnable projection $\overline{\epsilon}$, for Nystr\"omformer the number of selected landmarks and for Performer the number of random features.
While in principle scale could be defined as a constant, instead of a proportion of the input length (i.e, $DCT-0.25$ for $n=128$ implies $\overline{n}=32$), we argue that it would be mathematically unfounded to assume that is possible to obtain a constant complexity for an arbitrary input length, whatever efficient attention head is used. \melo{From the reported results it is clear that the transformer model, equipped with our DCT based efficient attention, outperforms all the competitors. As expected and discussed in \Cref{sec1:2} the savings in memory and inference times, from the usage of our attention head, are directly proportional to the sequence length. In the next paragraph we discuss this important aspect in more details.}

\begin{figure}[h!]
    \centering
    \includegraphics[clip, width=1.0\textwidth]{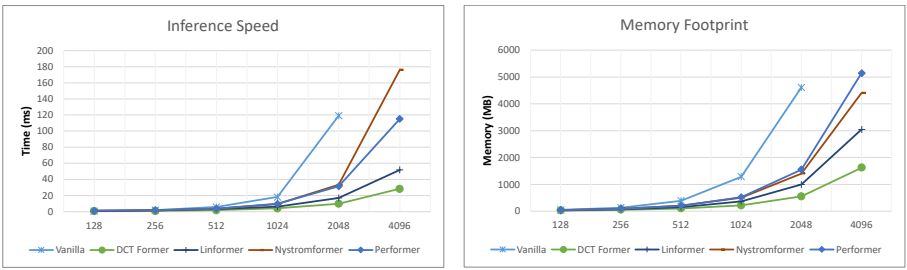}
    \caption{Plot of Inference Time (ms) and Memory Footprint (MB) for input sequences of different lengths. The Batch Size is fixed to 16 to allow for comparison.}
    \label{fig:fixedbs}
\end{figure}

\melo{\paragraph{Scalability with Sequence Length}}
\melo{
To obtain the inference results presented in the last paragraph we were forced to reduce the batch-size (BS) when increasing the sequence length (N) in order to fit the model in memory. While this approach is perfectly suitable to compare different models - for a fixed sequence length - it does not allow to truly appreciate how each model scale with the sequence length. We setup a new experiment to evaluate the growth in memory occupation and inference times when we vary the sequence lengths. In fact, as reported in \Cref{fig:fixedbs} we benchmark exclusively the multi-head attention modules with a small fixed batch size. For this experiment we maintain, for each attention, the same scale factors of \cref{Table1}.
}

\melo{
\noindent As deducible from its formulation, the vanilla attention scales proportionally with the square of the input's length. Our efficient attention outperforms all the competitors in terms of scalability, both in inference times and memory footprint. In particular, for the longest input sequences the benefit of the efficient attention reflects in a memory reduction of up to $80\%$, thus successfully enabling to work with significantly longer sequences.\\
In the following section we instead present the results for the trained models on both the pretraining and downstream tasks, showing that our attention can perform competitively when compared to significantly heavier formulations.
}

\subsection{Results and Discussion}\label{sec4:3}
We evaluate multiple settings of our model following the configurations detailed in Section \ref{sec4:1}. For the pretraining stage, we report both the best loss (Cross Entropy) on the validation set and the Accuracy score for the MLM task. It is worth remembering that the MLM can be evaluated as a multilabel classification problem, since for each masked token of the sentence we aim at predicting the correct vocabulary entry index (which in our case is $30522$ entries long). In addition, to make the comparison fair we evaluate the \textit{normalized} accuracy score, \melo{which is obtained by dividing the accuracy score by the normalized average inference time obtained by the same model with a bath size of 256 and a sequence length of 128 (divided by $10^2$ for readability).} For the finetuning we report the averaged Precision, Recall and F1-score obtained on the test split.\\
\begin{table}[h!]
\begin{center}
\caption{Results of transformer models with different attentions}
\label{Table2}
\begin{tabular}{ccccccc} 
\toprule
\begin{tabular}[c]{@{}c@{}}\\\\\\\end{tabular} & \multicolumn{3}{c}{Pretraining}                                      & \multicolumn{3}{c}{Finetuning $\uparrow$}  \\ 
\cmidrule{2-7}
Attention                                      & Loss $\downarrow$ & Accuracy (\%) $\uparrow$ & Normalized $\uparrow$ & Precision & Recall & F1-Score              \\ 
\midrule
Vanilla                                        & 2.07                 & 59.7                     & 1.52                  & 0.9         & 0.9      & 0.9                     \\
DCT-16                                         & 2.58                 & 51.6                     & 1.73                  & -         & -      & -                     \\
DCT-32                                         & 2.36                 & 54.7                     & \textbf{1.74}         & 0.87         & 0.87      & 0.87                     \\
IDEAL-32                                       & 2.26                 & 56.6                     & -                     & 0.88         & 0.88      & 0.88                     \\
DCT-48                                         & 2.28                 & 56.0                        & 1.68                     & 0.86         & 0.85      & 0.85                     \\
DCT-64                                         & 2.24                 & 56.6                        & 1.61                     & 0.85         & 0.85      & 0.85                     \\ 
\cmidrule(lr){1-7}
Linformer-16                                   & 2.29                 & 56.2                     & 1.49                   & 0.80         & 0.80      & 0.80                     \\
Linformer–32                                   & 2.17                 & 57.9                        & 1.49                     & 0.82         & 0.82      & 0.82                     \\
Linformer–48                                   & 2.13                 & 58.5                        & 1.49                     & 0.83        & 0.83 & 0.83                     \\
Nystrom–16                                     & 2.25                 & 56.6                        & 1.37                     & 0.88         & 0.87      & 0.87                     \\
Nystrom–32                                     & 2.13                 & 58.8                        & 1.26                     & 0.88         & 0.88      & 0.88                     \\
\bottomrule
\end{tabular}
\end{center}
\footnotetext{For all the results the number of DCT coefficients (and respectively of Nystr\"om landmarks and Linformer $\overline{\epsilon}$ size) is fixed. We use the notation $\mathbf{\{Model\}-\{\overline{n}\}}$}
\end{table}
All the sequences of the pretraining data are close to 128 tokens, requiring only a minimal amount of padding to be fixed to exactly 128 tokens: the finetuning sequences instead, while being truncated to 1024 tokens, presents a significant length variation.
For this reason we used the Makhoul's method to compute the forward and inverse DCTs in the finetuning phase. With respect to the usage of the $\overline{D}$ matrix, this allows us to work with any sequence length, while otherwise we would be limited to only work with sequences of exactly $\overline{n}$ tokens. Finetuning the Linformer it is instead far more problematic: the matrices $\overline{\epsilon}_{i}$ learned during the pretraining are only suitable to work for sequences of $128$, the only way to perform the finetuning is hence to reinitialize the transformation matrix to work with sequences of $1024$ and zero-pad all dataset elements to the maximum length. With Nystr\"omformer we encountered a similar issue, since also in this case the sequence length is required to be known and be evenly divisible by the number of landmarks, hence we opted to pad the sequences in the same way as for Linformer.\\
For the results reported in Table \ref{Table2}, the models trained with the efficient attention formulation of Algorithm \ref{algo1} are reported as DCT-$\{\overline{n}\}$, while the experiment IDEAL-32 follow the formalization of Algorithm \ref{algo2} to evaluate the approximation error induced by the DCT compression, without leveraging the relaxation on the softmax operation. It is fundamental to understand that, while the Ideal setup clearly outperforms the efficient setup, the Ideal setup needs to compute all the matrices onto the $\RR^n$ space, therefore losing all relevance to both memory and speed efficiency. Exploring alternatives to our softmax relaxation is a potentially interesting topic on its own, and can represent a future research direction.\\.


\section{Conclusion}\label{sec5}

In this work we analyzed the transformer architecture, in particular we focused on the attention mechanism, which grows for memory and computational time quadratically in the input length. Since in practical applications we are potentially faced with text sequences of thousands of words or videos of hundreds of frames, this growth represents the real bottleneck of these architectures. Our method, on the other hand, allows choosing the size $\overline{n}$ of the workspace, compressing the available information through the DCT. Once we have set a compression threshold, in line with our competitors, the experiments carried out show that our method requires a memory allocation that is a quarter less than the standard attention and saves a fifth of the inference time, while still maintaining a comparable expressive capacity. Due to its great flexibility, we consider the proposed method particularly suitable for all the applications with large amounts of data. In fact, contrary to other approximations proposed in the literature, our method allows for greater adaptability and ease of applicability, by not requiring the length of the sequence to be known in advance. Then the desired memory and time usage can be chosen by defining the number of DCT coefficients to be used. As a final reminder, energy-efficiency represents a raising concern for large deep learning models: reducing inference and training cost represents one of the biggest challenges for the near future. We are confident that our work could inspire other researchers in the domain of GreenAI.
\backmatter
\bmhead{Acknowledgments}
This work has been partially supported by the INdAM research group GNCS.

\begin{appendices}
\end{appendices}

\pagebreak


\bibliography{bibliography}

\end{document}